\documentclass{article}
\usepackage{ijcai26}

\usepackage{times}
\usepackage{soul}
\usepackage{url}
\usepackage[hidelinks]{hyperref}
\usepackage[utf8]{inputenc}
\usepackage[small]{caption}
\usepackage{graphicx}
\usepackage{amsmath}
\usepackage{amsthm}
\usepackage{booktabs}
\usepackage{algorithm}
\usepackage{algorithmic}
\usepackage[switch]{lineno}

\title{DataCube: A Video Retrieval Platform via Natural Language Semantic Profiling}

\author{
Yiming Ju$^*$
\and
Hanyu Zhao$^*$
\and
Quanyue	Ma
\and
Donglin	Hao
\and
Chengwei Wu\and
Ming	Li\and
Songjing	Wang\and
Tengfei	Pan$^{\dag}$\\
\affiliations
Beijing Academy of Artificial Intelligence\\
\emails
\{ymju, hyzhao, tfpan\}@baai.ac.cn
}

\begin{document}

\maketitle
\renewcommand{\thefootnote}{}
\footnotetext{\textsuperscript{*}Equal Contribution. \textsuperscript{\dag}Corresponding Author.}
\renewcommand{\thefootnote}{\arabic{footnote}}

\begin{abstract}

Large-scale video repositories are increasingly available for modern video understanding and generation tasks. However, transforming raw videos into high-quality, task-specific datasets remains costly and inefficient. We present DataCube, an intelligent platform for automatic video processing, multi-dimensional profiling, and query-driven retrieval. DataCube constructs structured semantic representations of video clips and supports hybrid retrieval with neural re-ranking and deep semantic matching. Through an interactive web interface, users can efficiently construct customized video subsets from massive repositories for training, analysis, and evaluation, and build searchable systems over their own private video collections. The system is publicly accessible at \url{https://datacube.baai.ac.cn/}. \textbf{Demo Video:} \href{https://baai-data-cube.ks3-cn-beijing.ksyuncs.com/custom/Adobe\%20Express\%20-\%202\%E6\%9C\%8818\%E6\%97\%A5\%20\%281\%29\%281\%29\%20\%281\%29.mp4}{\underline{Click here} to view the demo video.}

\end{abstract}


\section{Introduction}

The rapid development of video generation and understanding models has raised higher demands on data quality, composition, and customization \cite{zhou2024survey}. Although large-scale open-domain video repositories and open-source datasets at the petabyte scale are increasingly available, such as HowTo100M \cite{miech2019howto100m}, Panda-70M \cite{chen2024panda}, Koala-36M \cite{wang2025koala}, and several recent large-scale video datasets 
\cite{nan2024openvid,ju2025miradata,wang2023internvid,chen2024sharegpt4v,ju2025cividcoherentinterleavedtextvideo}.
Most of them cannot be directly utilized due to the difficulty of extracting task-specific content and controlling dataset distributions. As a result, constructing customized video datasets remains costly and time-consuming.

Previous works have explored CLIP-based embedding methods for indexing large-scale video repositories and enabling similarity-based retrieval \cite{luo2022clip4clip,portillo2021straightforward,nguyen2024videoclip,zhao2022centerclip}. While effective and efficient in practice, these approaches typically rely on a single global embedding to capture overall visual--textual similarity. Such representations offer limited flexibility for modeling fine-grained semantics and make it difficult to extend the system once the indexing pipeline is established.

Recent advances in multimodal understanding models have made it possible to transform video content into structured natural-language representations~\cite{nwae403}. Based on such representations, videos can be described as collections of interpretable semantic attributes, enabling semantic retrieval and analysis. Compared with CLIP-based similarity retrieval methods, this paradigm is more scalable, as semantic annotations can be flexibly extended to multiple attributes, and more interpretable, facilitating error diagnosis and semantic tracing.
By leveraging pre-computed semantic representations over massive video repositories, such semantic-aware retrieval systems enable large-scale video understanding and personalized data access. Users can efficiently construct customized datasets without repeatedly reprocessing the entire corpus, thereby avoiding redundant computation and unnecessary resource consumption.


In this work, we present DataCube, an end-to-end intelligent platform for video processing, profiling, and semantic retrieval. DataCube integrates automatic preprocessing, multi-dimensional semantic modeling, and hybrid retrieval mechanisms to enable efficient construction of customized video datasets from large-scale repositories. By transforming raw videos into structured natural-language semantic representations, DataCube supports flexible and personalized video retrieval based on content, style, and other semantic constraints. A coarse-to-fine retrieval framework is adopted to balance efficiency and accuracy, combining embedding-based filtering with deep semantic matching for complex queries.

Built upon large-scale open datasets and continuously expanded through user-contributed repositories, DataCube maintains a searchable index of hundreds of millions of video clips. The platform is publicly accessible through a web interface at \url{https://datacube.baai.ac.cn/}, supports both Chinese and English queries, and has attracted over one thousand registered users. In addition, it has been adopted in collaborative research projects with multiple universities.

\begin{figure*}[t!]
  \centering
   \includegraphics[width=1\linewidth]{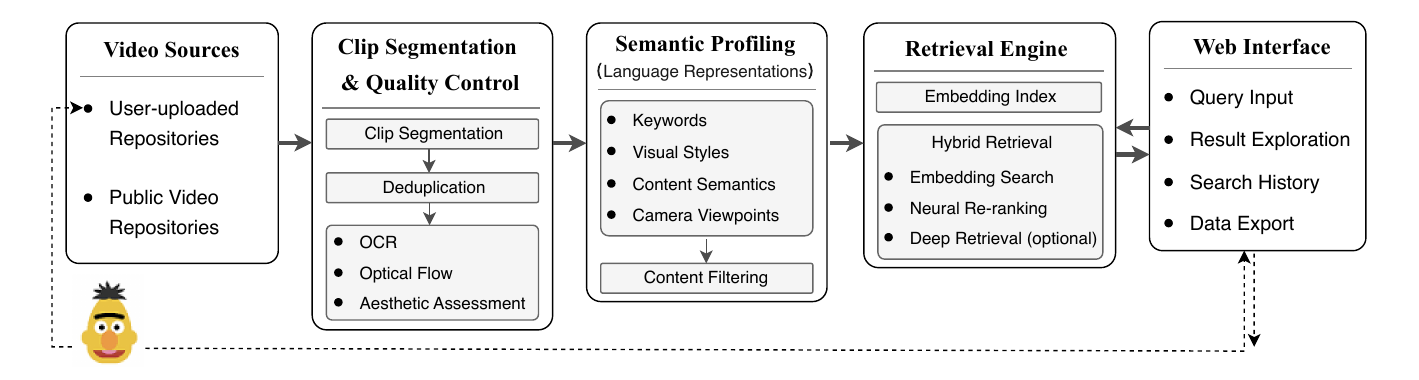}
   \caption{System Overview.}
   \label{ijcai1}
\end{figure*}

\section{System Overview}

DataCube is an end-to-end platform for automatic video processing, profiling, and semantic retrieval, which is built upon large-scale open datasets, including Runway and Koala~\cite{wang2025koala} collections, and supports data ingestion from user-uploaded private repositories. As illustrated in Figure~\ref{ijcai1}, the system transforms large-scale raw video repositories into profile-enriched and searchable databases through a unified pipeline, enabling customized retrieval and dataset construction.

As shown in Figure~\ref{ijcai1}, video sources are first automatically segmented into clips and processed by a multi-stage quality control pipeline, including deduplication, optical flow analysis, OCR detection, and aesthetic assessment.
Next, DataCube constructs multi-dimensional semantic profiles in natural language using large vision-language models (VLMs). These profiles capture visual styles, content semantics, camera viewpoints, and associated keywords, and are further used to filter potentially non-compliant videos. The remaining profiles are then encoded into dense embeddings for efficient indexing.

On the retrieval side, user queries are processed by a hybrid retrieval engine. Queries are first enriched and analyzed, and then mapped to different profiling dimensions. The mapped representations are used for embedding-based retrieval and neural re-ranking. A deep retrieval mode is further provided for complex queries, enabling direct video-query comparison for high-precision filtering.
Finally, DataCube exposes all functionalities through an interactive web interface, allowing users to explore retrieval results, browse historical search records, and export customized video subsets.

\begin{figure*}[t!]
  \centering
   \includegraphics[width=1\linewidth]{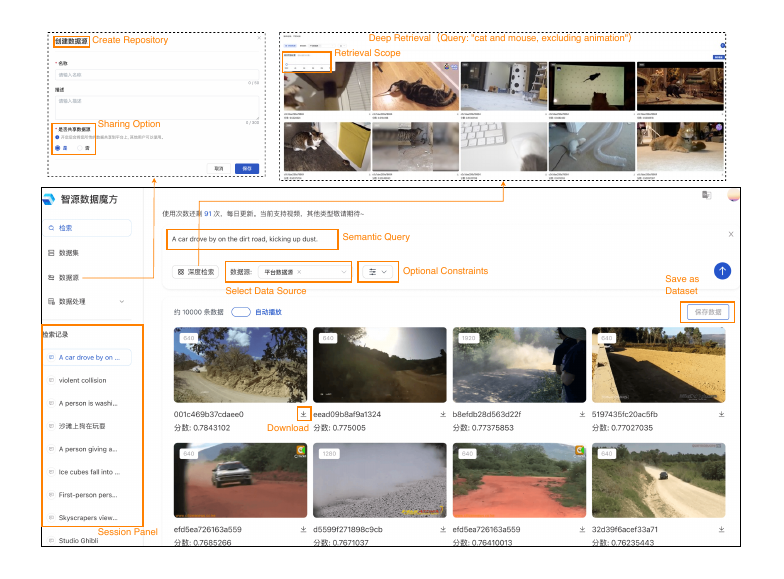}
   \caption{Demonstration Scenario of DataCube.}
   \label{ijcai2}
\end{figure*}

\section{Key Components and Implementation}

This section outlines the key implementation of DataCube.

\paragraph{Data Storage and Preprocessing.}
All videos and generated clips are stored in KS3 object storage. Raw videos are segmented into semantic clips using content-aware scene detection with PySceneDetect\footnote{\url{https://github.com/Breakthrough/PySceneDetect}} (threshold = 26). Clips shorter than five seconds are discarded, while raw videos are archived in cold storage.
During preprocessing, DataCube applies URL-based and frame-hash-based deduplication. Text coverage is estimated using PaddleOCR\footnote{\url{https://github.com/PaddlePaddle/PaddleOCR}}, motion strength is measured by computing optical flow every 0.5 seconds with RAFT~\cite{teed2020raft}, and clip-level aesthetic quality is estimated using NIQE~\cite{niqe} and MUSIQ~\cite{musiq}. Based on these signals, nearly static clips (motion score $<10$) are filtered out, while other attributes are retained as user-configurable filtering options.

\paragraph{Semantic Profiling and Indexing.}
Qualified clips are analyzed using Qwen2.5-VL-7B~\cite{wang2024qwen2} to construct natural-language semantic profiles, covering keywords, visual styles, content semantics, and camera viewpoints. Each clip is represented by a concatenated image composed of uniformly sampled frames. The resulting profiles are encoded into dense embeddings using BGE\footnote{\url{https://huggingface.co/BAAI/bge-large-en-v1.5}} and indexed with Milvus. CPU and GPU workloads are dynamically scheduled using Ray, enabling scalable indexing over hundreds of millions of clips.

\paragraph{Hybrid and Deep Retrieval.}
User queries are first enriched using a GPT-based interface and encoded with BGE embeddings to retrieve candidate subsets from Milvus (default size: 10{,}000). The retrieved candidates are then re-ranked using Qwen3-Reranker-0.6B~\cite{zhang2025qwen3}. For high-precision requirements, Qwen2.5-VL-72B~\cite{wang2024qwen2} performs direct video-query semantic matching for deep retrieval. The deep retrieval module is deployed with vLLM~\cite{kwon2025vllm} on A100 GPUs and supports long-running background inference tasks.

\section{Demonstration Scenario}

DataCube exposes all functionalities through an interactive web interface, allowing users to efficiently explore retrieval results, browse historical search records, and export customized video subsets. Figure~\ref{ijcai2} illustrates the overall demonstration workflow.

\paragraph{Interactive Query and Retrieval.}

Users can \textbf{select a data source} for retrieval, either from the public platform repository or from private repositories\footnote{Users can upload raw videos to DataCube, which automatically processes the data and constructs an indexed repository. Once processing is completed, the repository becomes selectable for retrieval. DataCube supports retrieval from multiple data sources, which can be freely combined within a single query.
}. When creating a repository, users may choose whether to \textbf{share it with other users} or keep it private. 
During retrieval, users specify \textbf{semantic queries} in natural language and configure \textbf{optional constraints} through a control panel, such as target resolution and clip duration ranges.
After submission, DataCube executes the hybrid retrieval pipeline and presents ranked results in a grid layout, where each video is displayed together with its relevance score. Retrieved results are simultaneously recorded in a \textbf{session panel} on the left sidebar, enabling users to track multiple retrieval tasks.

\paragraph{Result Management and Dataset Export.}
Individual videos can be \textbf{previewed and downloaded} directly through the web interface. Users may further select a subset of the top-ranked retrieved videos (up to 10{,}000) and \textbf{save them as a dataset}, upon which the system automatically packages the selected samples and generates a downloadable link for offline usage.

\paragraph{Deep Retrieval for Complex Queries.}
Users can enable the \textbf{deep retrieval} mode. In this mode, DataCube launches background matching tasks and records their execution status in the sidebar. Due to the increased computational cost, this mode processes up to 500 candidate videos by default and typically completes within 3--5 minutes. Users may optionally expand the \textbf{retrieval scope} to 10{,}000 candidates, with processing time increasing approximately linearly.
Compared with standard retrieval, deep retrieval provides more accurate semantic matching and supports complex constraints. For example, Figure~\ref{ijcai2} presents retrieval results for a query that includes an explicit exclusion requirement.
Once the task is completed, refined results can be accessed from the sidebar and downloaded as a filtered dataset.

\section{Conclusion}

We introduce DataCube, an end-to-end platform for large-scale video processing and semantic retrieval. By combining structured semantic profiling with hybrid and deep retrieval mechanisms, DataCube enables efficient construction of customized video datasets from massive repositories through an interactive web interface. The system demonstrates the practicality of semantic-aware video retrieval by promoting data sharing and semantic profile reuse, thereby effectively reducing data preparation costs and supporting video-centric research and applications.
















\bibliographystyle{named}
\bibliography{ijcai26}

\end{document}